\documentclass[journal]{IEEEtran}
\usepackage{graphicx}
\usepackage{url}
\usepackage[nocompress]{cite}
\usepackage{array}

\begin{document}

\title{Human Modelling and Pose Estimation Overview}

\author{\IEEEauthorblockN{Pawel Knap\IEEEauthorrefmark{1}}
\IEEEauthorblockA{\\University of Southampton\\
\IEEEauthorrefmark{1}pmk1g20@soton.ac.uk}}

\markboth{ArviX, \today}%
{Knap: Human Modelling Overview}

\maketitle

\begin{abstract}
Human modelling and pose estimation stands at the crossroads of Computer Vision, Computer Graphics, and Machine Learning. This paper presents a thorough investigation of this interdisciplinary field, examining various algorithms, methodologies, and practical applications. It explores the diverse range of sensor technologies relevant to this domain and delves into a wide array of application areas. Additionally, we discuss the challenges and advancements in 2D and 3D human modelling methodologies, along with popular datasets, metrics, and future research directions. The main contribution of this paper lies in its up-to-date comparison of state-of-the-art (SOTA) human pose estimation algorithms in both 2D and 3D domains. By providing this comprehensive overview, the paper aims to enhance understanding of 3D human modelling and pose estimation, offering insights into current SOTA achievements, challenges, and future prospects within the field.
\end{abstract}

%\begin{IEEEkeywords}
%\end{IEEEkeywords}

\IEEEpeerreviewmaketitle

\section{Introduction}
Human modelling encompasses a range of techniques that include human pose estimation (HPE) and visualization of 3D human models. Situated at the intersection of Computer Vision (CV), Computer Graphics (CG), and Machine Learning (ML), this field integrates various methodologies to create accurate representations of human anatomy in 3D space. In our study, we delve into different algorithms employed for these tasks and conduct a comparative analysis of state-of-the-art (SOTA) solutions.

HPE involves identifying the poses of human body parts and joints, usually in images or videos. Various sensors can be employed for this task, each offering distinct advantages and drawbacks. Monocular cameras \cite{XNect_SIGGRAPH2020, pishchulin2016deepcut, insafutdinov2016deepercut, pifpaf}, commonly utilized due to their affordability, are hindered by limitations such as occlusion and depth ambiguity. Camera arrays or multiple camera motion capture systems \cite{Human3.6M} mitigate some monocular camera limitations but are costlier and less versatile. RADAR systems \cite{adib_capturing_2015,sengupta_nlp_2020,sengupta_mm-pose_2020-1,an_fast_2022,li_capturing_2020,xue_mmmesh_2021,huang_indoor_2021,lee_hupr_2022} excel in occlusion handling and privacy preservation, yet encounter challenges due to sparse data input. Similarly, LIDAR sensors \cite{6, ye_lpformer_2023} face sparse data issues while also being costly; however, they offer high-resolution output. Infrared (IR) sensors like Kinect \cite{7, kinect} face difficulties outdoors due to sunlight interference. Wearable Motion Capture Systems offer another option for HPE but are constrained by costs and intrusiveness, restricting their applications. Moreover, there exist hybrid systems utilizing a combination of the aforementioned sensors \cite{previous, Knap_2023, Knap}. Therefore, our focus in this study predominantly centres on camera-based solutions, as the current research direction is to overcome their limitations, positioning them as SOTA solutions. Works involving RADAR, LIDAR, or IR-based approaches are comparatively less advanced. %However, we will also briefly mention them.

\begin{figure}[ht!]
    \includegraphics[width=0.5\textwidth]{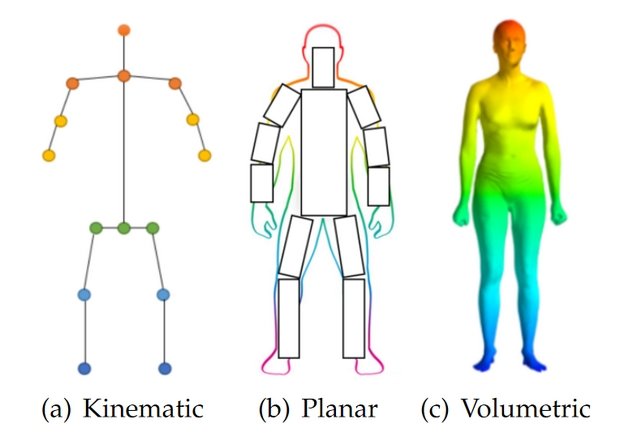}
    \caption{Different representations of humans used in HPE system's visualisations. Source \cite{zheng2023deep}}
    \label{pic: types of pose}
\end{figure}

Once the poses are identified, CG techniques are employed to visualize these real-world poses on a computer screen. Various representations of the human body, as depicted in Figure \ref{pic: types of pose}, can be used. These representations can be further manipulated, altering pose shape or appearance. Additionally, they can be animated, exemplifying their usefulness for the movie or gaming industry. Other application domains include virtual (VR), and augmented reality (AR) \cite{2023,5597097}, which currently stand as the most actively explored scientific research avenues for HPE.

\begin{figure*}
    \includegraphics[width=1\textwidth]{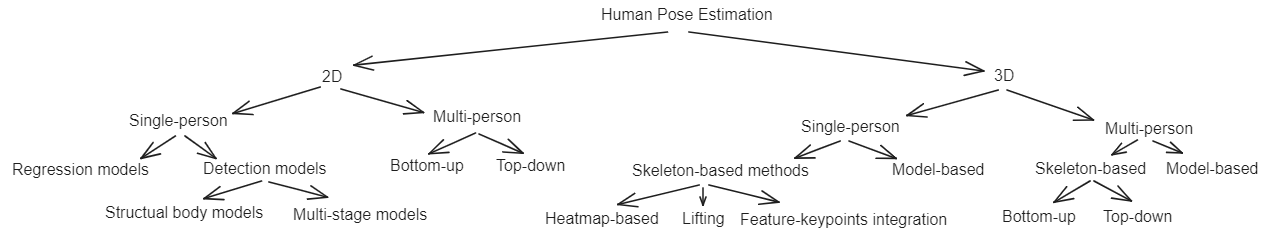}
    \caption{Dichotomy of Human Pose Estimation algorithms.}
    \label{pic: dichotomy}
\end{figure*}

HPE can also be used in Human-Computer Interactions (HCI) \cite{8411908,9498024}, for gesture control, improving user interaction with digital devices. In Robotics \cite{Vasileiadis_2017_ICCV,a,b}, HPE improves robots' ability to intuitively interact with humans, particularly in human assistant roles. It also plays a crucial role in video surveillance \cite{LAMAS2022488,Cormier_2021_ICCV,Cormier_2022_WACV}, aiding in the recognition of suspicious actions, and automotive industry for self-driving cars \cite{self_drivingHPE2023, slef-drivingHPE,pifpaf}. Furthermore, in sports or rehabilitation \cite{1,dd,2021}, HPE is utilized for analyzing movements to evaluate performance and enhance training techniques. Similarly, it can recognize posture defects for healthcare applications. Additionally, HPE is a key step towards the concept of digital twins \cite{digitaltwin_2020,digitaltwin_2021}, which would be useful in designing personalized disease treatments or user-focused architecture.

The rest of the paper is organized as follows: Section 2 offers background information, while Section 3 reviews popular datasets and metrics. Section 4 delves into cutting-edge algorithms, comparing them, and outlining the problems they address, advantages, disadvantages, specifics, and target applications. Section 5 provides a comprehensive discussion of future research directions, and the paper concludes in Section 6.

\section{Background}
Constrained by limited data and computational resources, early HPE research was concentrated on crafting features manually or optimizing deformable human body models. However focus changed due to the recent advancements in deep learning (DL) techniques, and in the availability of large-scale 2D/3D pose datasets such as COCO \cite{COCO}, MPII \cite{MPII}, Human3.6M \cite{Human3.6M}, and 3DPW \cite{von2018recovering}. DL techniques have partially addressed challenges like occlusion, close inter-person interactions, crowded scenes, complex postures, and small target persons. Neural networks (NN) automatically learn crucial features and capture non-linear information, offering advantages over previous methods based on hand-crafted features. However, they also have drawbacks, including sensitivity to small image changes \cite{visual_instability_1}, difficulties in generalization \cite{generalisation}, and issues with explainability and interoperability \cite{interpret}.

There are diverse methods for representing the detected human body parts. These include keypoint-based representations in 2D or 3D, or 2D/3D heatmaps exhibiting high responses at keypoint locations and being suitable for regression by NN. Different methods use orientation maps like Part Affinity Fields (PAF) \cite{openpose}, which utilize a 2D vector field of each pixel pointing from one limb to the other. This concept was further extended to 3D \cite{orinet}. Another keypoint-based approach is 2D compositional human pose (CHP) \cite{CHP}, which introduces a blend of bone and limb vectors and has been further developed into 3D \cite{CHP_3D,CHP_3D_2}. Alternatively, richer information can be obtained using model-based representations. These are categorized into part-based volumetric models (planar models), representing limbs as geometric figures like cylinders \cite{planar_cylinder} or ellipsoids \cite{planar_ellipses}, and detailed statistical 3D human body models like the skinned multi-person linear model (SMPL) \cite{SMPL}, encoding shape and pose into low-dimensional parameters.

Nowadays, human modelling and pose estimation algorithms can be categorized into 2D and 3D approaches, and further divided based on whether they focus on single or multiple subjects. In our study, we will explore algorithms from these four distinct groups as shown in figure \ref{pic: dichotomy}.

\subsection{2D Single-Person Human Pose Estimation}
\begin{figure}[ht!]
    \includegraphics[width=0.48\textwidth]{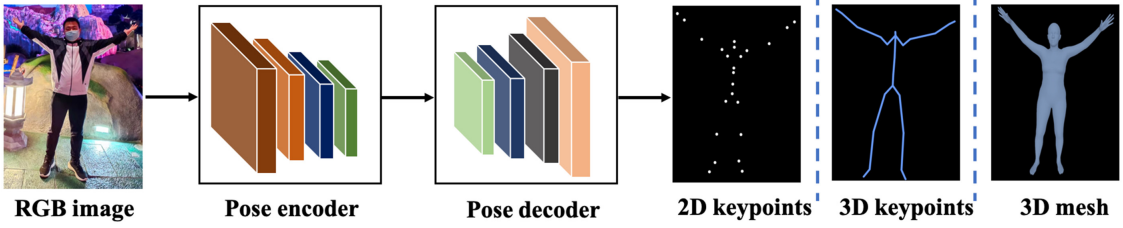}
    \caption{Standard framework for estimating the pose of a single individual. Image taken from \cite{survey_HPE}.}
    \label{pic: single HPE}
\end{figure}

The typical architecture for single-person HPE depicted in Figure \ref{pic: single HPE} consists of a pose encoder and decoder. The encoder extracts features that are used by the decoder to determine human keypoints through regression or detection. While pre-trained models such as ResNet \cite{ResNet} are commonly employed as encoders, there is a growing interest in specialized pose-focused encoders like the hourglass network \cite{hourglass19} and PoseNAS \cite{NAS}. These architectures leverage techniques such as skip connection layers and neural architecture search (NAS) to enhance feature extraction for pose-related tasks.

However, the major focus is put on the design of the decoder. Regression models, such as DeepPose \cite{deeppose}, which is a first method using deep convolution neural networks (CNN), Iterative Error Feedback (IEF) network \cite{carreira2016human} using self-correcting model, and Sun et al. \cite{CHP} model utilizing body structure-aware compositional pose regression, have been milestones. However, regression models often struggle with complex poses due to the inherent non-linearity of directly mapping images to body coordinates. 

An alternative approach involves a detection-based decoder using heatmaps, which can be categorized into two main types: multi-stage architectures and structural body models. Structural body models leverage anatomical knowledge; for instance, Tompson et al. \cite{tompson2014joint} use a Markov Random Field to represent body joint distributions, Chen et al. \cite{chen2014articulated} employ conditional probabilities to learn location and spatial relationships of body part, and Chen et al. \cite{chen2017adversarial} utilize a structure-aware network. On the other hand, multi-stage architectures extract features for HPE at different levels. The well-known Hourglass network \cite{hourglass19} and its variants, such as \cite{yang2017learning}, which combines the Hourglass network with a feature pyramid, and \cite{chu2017multi}, which introduces Hourglass Residual Units to increase the network's receptive field, fall under this category. Other notable architectures include the Convolutional Pose Machine (CPM) \cite{wei2016convolutional}, which utilizes intermediate inputs and lacks an explicit graphical model.

Moreover, active research focuses on multi-task learning and pose refinement techniques. This includes joint 2D and 3D HPE \cite{luvizon20182d} and iterative pose refinement modules \cite{carreira2016human}. Additionally, efforts to enhance speed and efficiency, particularly for video applications, involve lightweight networks \cite{rafi2016efficient,debnath2018adapting}, network binarization \cite{bulat2017binarized}, and model distillation \cite{li2021online,zhang2019fast}. Using temporal information across frames in video processing can enhance HPE results. Approaches such as optical flow-based feature extraction \cite{pfister2015flowing,song2017thin} and sequence models like recurrent networks \cite{gkioxari2016chained} and LSTM \cite{luo2018lstm} have shown promise in this domain.

\subsection{2D Multi-Person Human Pose Estimation}
\begin{figure}[ht!]
    \includegraphics[width=0.48\textwidth]{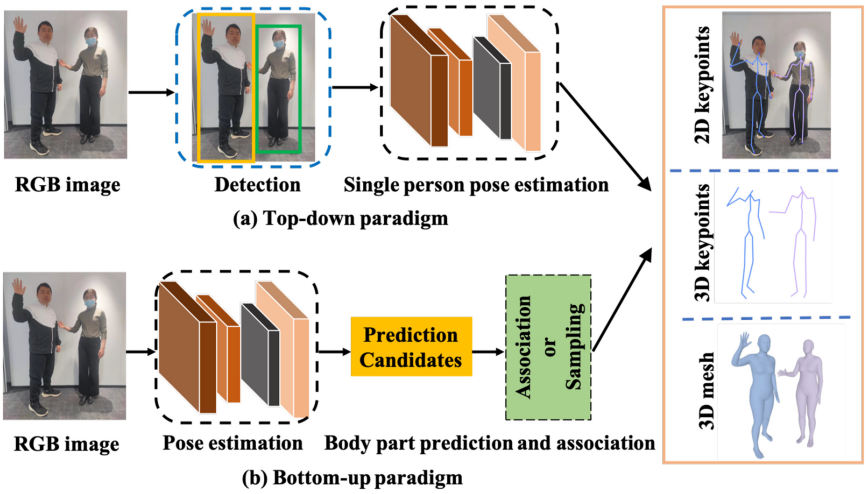}
    \caption{Standard frameworks for estimating poses of multiple people. Both top-down and bottom-up methods use encoder-decoder architecture. Image taken from \cite{survey_HPE}.}
    \label{pic: multi HPE}
\end{figure}

\begin{figure*}
    \includegraphics[width=1\textwidth]{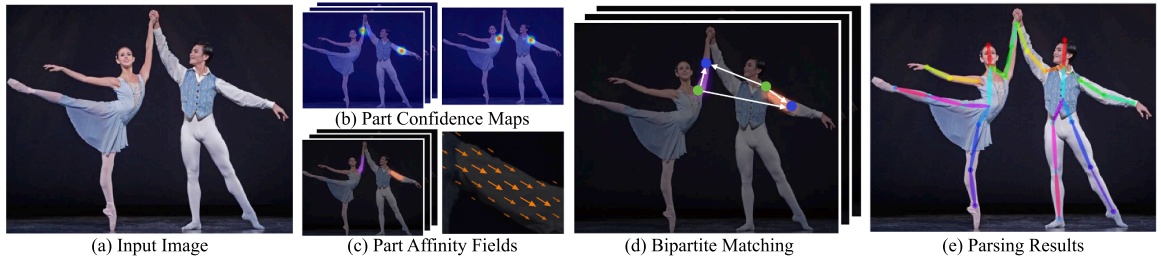}
    \caption{The workflow of the bottom-up approach utilized in OpenPose \cite{openpose}. The (b) Part Confidence Maps represent the heatmaps of body parts. Following the prediction of (c) Part Affinity Fields, (d) Bipartite Matching is executed to correlate body part candidates, culminating in the derivation of the (e) Parsing Results. Image source: \cite{openpose}.}
    \label{pic: openpose}
\end{figure*}

The multi-person HPE estimation is significantly more difficult than the single-person one because, in a multi-person scenario, an algorithm has to be able to detect an unknown number of people in sometimes crowded scenes, where inter-person or object occlusion happens often, and small targets occur. For this group of algorithms, we can distinguish two methods shown in figure \ref{pic: multi HPE}: bottom-up, where body parts of people are firstly detected and subsequently these parts are combined together for each person. And top-down, where each person is firstly localised and then in the bounding box of such person its body parts are detected, often for that purpose single-person HPE algorithms are used.

\subsubsection{Top-down methods}
However, single-person HPE algorithms are adjusted for the multi-person scenario to handle challenges such as truncation, occlusions, and small targets. Moreover, post-processing techniques like Non-Maximum-Suppression (NMS) \cite{papandreou2017towards} are applied to enhance keypoint detection after single-person HPE.

One of the pioneering top-down approaches was proposed by Papandreou et al. \cite{first_top_down}, which utilized a Faster RCNN detector and ResNet \cite{ResNet} to predict keypoint heatmaps. They also introduced keypoint-based NMS for improved localization. Xiao et al. \cite{xiao2018simple} aimed to enhance spatial resolution by incorporating three deconvolution layers into a ResNet backbone. Another significant advancement was made by Chen et al. \cite{COCO_2017} with the Cascade Pyramid Network (CPN), which combines a global network with subsequent refinement to improve keypoint prediction. They also introduced an online hard keypoints mining (OHKM) loss to address challenging keypoints. Another approach for difficult keypoints \cite{sun2019deep} maintains high-resolution images throughout the network and integrates high-to-low resolution sub-networks, creating multi-resolution features. The two above models fall into the category of multi-stage or multi-branch fusion models, which provide superior detection accuracy due to increased complexity.

A common issue in top-down models is the occurrence of multiple individuals within a single bounding box. To address this, RMPE \cite{rmpe} employs parametric pose NMS to eliminate redundant people. Additionally, it introduces a symmetric spatial transformer for person identification and a pose-guided human proposal generator to enhance efficiency. Crowdpose \cite{crowdpose} adopts a different approach by utilizing a graph model to separate joint candidates. Additionally, the authors introduce a crowded HPE dataset in this paper, along with the Crowd Index, a metric designed to assess image crowding levels. Furthermore, similarly to single HPE, a multi-task approach can also be employed in top-down multi-person HPE \cite{liang2018look,xia2017joint}.

\subsubsection{Bottom-up Methods}
These methods can be categorized into two groups: those that detect and group keypoints for each person, and those that directly regress the position of each person's keypoints. In the latter group, which is less popular and well-known, pixel-wise methods like CornerNet \cite{cornerNet} and CenterNet \cite{centerNet} stand out as prominent examples.

In the first group, the core challenge lies in accurately grouping keypoints belonging to the same person. DeepCut \cite{pishchulin2016deepcut} addresses this by modelling grouping as integer linear programming (ILP), albeit with high computational costs. PAF \cite{openpose}, on the other hand, have emerged as SOTA solutions. These are 2D vector fields that encode the directional relationship between body parts. OpenPose \cite{openpose}, a real-time HPE algorithm, utilizes PAFs to connect confidence maps for each body part through a greedy algorithm as shown in the method framework in figure \ref{pic: openpose}. PifPaf \cite{pifpaf} extends this concept by using Part Intensity Field (PIF) for body part localization and PAF for association. The other idea is to group detected keypoints using embedding features or tags. Newell et al. \cite{newell2017associative} generates detection heatmaps and embedding tags for each body keypoint. HigherHRnet \cite{cheng2020higherhrnet} improves upon HRNet \cite{sun2019deep} by aggregating HRNet features in a pyramid and utilizing associative embeddings. Additionally, multi-task algorithms like PersonLab \cite{personlab} can be used to simultaneously detect body keypoints heatmaps and human segmentation maps.

\subsubsection{Top-down and bottom-up methods in videos}
Leveraging temporal information across consecutive frames can improve HPE performance. Development in this area was possible due to datasets like PoseTrack \cite{posetrack}, offering multi-person in-the-wild images for HPE and tracking. In top-down methods, temporal data can be used in two ways. Tracking-by-detection \cite{girdhar2018detect,wang2020combining}, also called clip-based technique, tracks individuals' bounding boxes across frames for subsequent single-person 2D video HPE based on cropped images. The other technique utilizes an optical flow pose similarity to associate poses across frames \cite{xiao2018simple}.

In bottom-up methods, the temporal pose association can be solved as a linear programming problem utilizing spatio-temporal graph \cite{insafutdinov2017arttrack,posetrack}, extending image-based multi-person HPE solutions \cite{pishchulin2016deepcut,openpose}. However, these methods are slow, hindering real-time applications. Therefore, \cite{poseflow} utilizes optical flow for online inter-frame pose association and tracking. Furthermore, concepts such as PAF \cite{openpose}, and associative embeddings \cite{newell2017associative} can be extended to Temporal Flow Fields \cite{raaj2019efficient,fabbri2018learning,doering2018jointflow} and spatio-temporal embedding \cite{jin2019multi} for keypoints matching between consecutive frames. 
\subsection{3D Single-Person Human Pose Estimation}
Monocular 3D HPE encounters similar challenges as 2D HPE, compounded by specific issues such as the scarcity of in-the-wild 3D datasets and inherent depth ambiguity. Despite these challenges, 3D representation offers advantages, capturing not only 3D location but also detailed shape and body texture. These aspects are leveraged in model-based 3D HPE, distinguishing it from skeleton-based 3D HPE, which lacks detailed appearance information.

\subsubsection{Skeleton-based methods}
\begin{figure}[ht!]
    \includegraphics[width=0.48\textwidth]{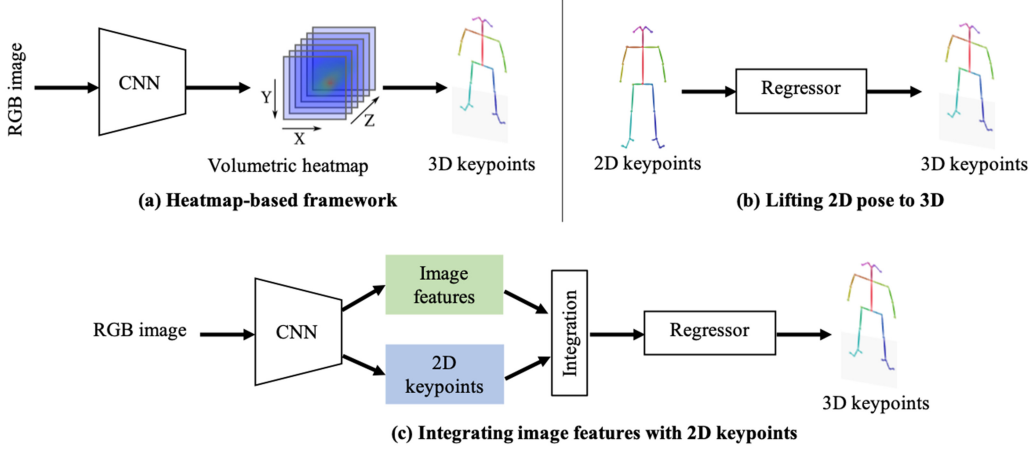}
    \caption{Exemplary frameworks for 3D single-person, skeleton-based HPE. Image taken from \cite{survey_HPE}.}
    \label{pic: 3d hope single}
\end{figure}

The skeleton-based 3D HPE can be divided into heatmap-based, 2D-3D lifting, and image features with 2D keypoints integration frameworks as shown in figure \ref{pic: 3d hope single}. In the first method, a CNN network maps each 3D keypoint in the image as a 3D Gaussian distribution in the heatmap. Then a postprocessing step finds a final keypoint location as a local maximum. For example, Pavlokas \cite{pavlakos2017coarse} introduces coarse-fine architecture consisting of hourglass networks \cite{hourglass19} to interactively get fine-grained keypoint location. VNect \cite{vnect} is a real-time video method that fits the kinematic skeleton in the post-processing to achieve temporally stable pose prediction. On the other hand, an integral method \cite{sun2018integral} can be used to directly get keypoints location from heatmaps.

2D-3D lifting methods start with stable 2D HPE algorithms followed by a depth-predicting network, thus utilizing 2D pose as an intermediate representation for 3D pose. However, accurately lifting the pose poses challenges due to inherent ambiguity — a single 2D pose can correspond to multiple 3D poses, and vice versa. Martinez et al. \cite{24} introduced a simple feedforward lifting NN that showed promising performance. Other methodologies, like those by Chen and Ramanan \cite{crowdpose} and Yang et al. \cite{Yang}, utilize exemplar-based techniques involving extensive dictionaries and nearest-neighbour searches for optimal 3D pose determination. Due to a lack of extensive datasets, weakly-supervised or unsupervised algorithms are often employed. The former methods use augmented 3D data or unpaired 2D-3D data to grasp human body priors without explicit 2D-3D correspondences. Notable contributions in this category include Pavlakos et al. \cite{pavlakos2018ordinal} and Ronchi et al. \cite{ronchi2018relative}, who integrated ordinal depth information into their methodologies. Additionally, \cite{wandt2019repnet} introduced a weakly-supervised adversarial method using kinematic chains.

Unsupervised 2D-3D lifting algorithms operate without labelled 3D pose data, relying on large-scale 2D annotated datasets and geometric constraints to infer 3D pose information. Tekin et al. \cite{tekin2016structured} proposed a structured prediction framework that jointly estimates 2D and 3D pose. In contrast, Tome et al. \cite{tome2017lifting} use a multi-stage CNN to refine the 2D pose using back projection, thus refining 3D estimations. Alternative lifting approaches include creating a library of corresponding 2D-3D poses by projecting 3D poses to 2D space using random cameras \cite{chen20173d}. Then, the predicted 2D pose is matched to a 2D pose from the library, and its 3D pair is obtained. However, integrating features extracted from the image with 2D keypoints simplifies 3D keypoints prediction by reducing depth ambiguity through contextual information. Thus, SemGCN \cite{zhao2019semantic240} employs joint-level features along with keypoints to construct graph nodes, enabling a graph CNN to predict 3D pose based on this information.

Due to the lack of 2D-3D datasets, other approaches utilize the multi-camera view and 2D pose dataset for supervision during training \cite{iqbal2020weakly}, or learn from synthetic data \cite{varol2017learning} which is created by projecting SMPL \cite{SMPL} or SCAPE \cite{SCAPE} 3D model onto 2D in-the-wild images. To further improve 3D skeleton model estimation, VideoPose3D \cite{pavllo20193d_156} uses temporal convolution on 2D poses, thus tackling the problem of inherent depth ambiguity, where a single 2D pose may correspond to multiple 3D poses.

\subsubsection{Model-based methods}
\begin{figure}[ht!]
    \includegraphics[width=0.48\textwidth]{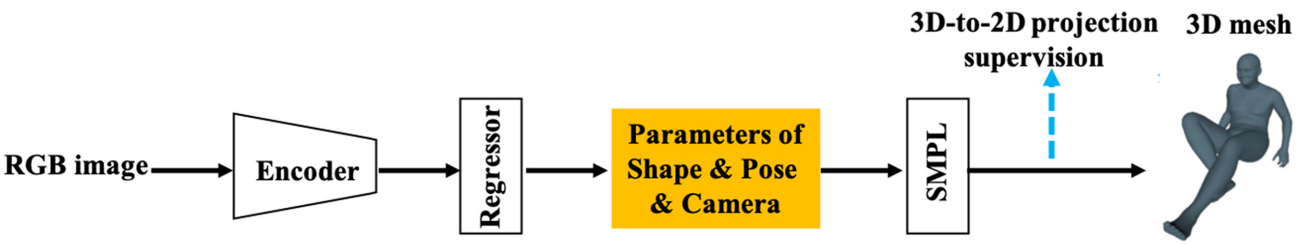}
    \caption{The exemplary framework for 3D HPE based on SMPL human model. Image taken from \cite{survey_HPE}.}
    \label{pic: SMPL}
\end{figure}

These methods are currently more popular than skeleton-based approaches as they offer finer details about human appearance, which is valuable for VR and AR applications. However, human representation are often simplified by utilizing statistical 3D human models like SMPL \cite{SMPL} instead of complex mesh regression techniques, thus streamlining the entire process. The representative framework for SMPL-based HPE algorithm is shown in figure \ref{pic: SMPL}.

Efforts have been made to explore alternative representations. For instance, GraphCMR \cite{kolotouros2019convolutional} uses a graph CNN to estimate SMPL parameters from nodes through regression, where each node represents a vertex of the SMPL template mesh and image feature vectors. Similarly, \cite{lin2021end} maps CNN features to joints and vertices of the mesh model to regress them to 3D coordinates using a transformer-based network. Other approaches extend SMPL and incorporate representations for the entire body. For example, \cite{romero2022embodied} integrates SMPL with a 3D hand model, while SMPL-X \cite{pavlakos2019expressive} further integrates the FLAME head model \cite{li2017learning}. \cite{pavlakos2019expressive} also introduce SMPLify-X, which iteratively fits SMPL-X to 2D keypoints of the body, face, and hands to generate the whole-body mesh.

Due to the limited availability of datasets, unpaired 2D-3D motion capture (MoCap) datasets are used to supervise the rationality of estimated SMPL pose parameters in a generative adversarial manner \cite{kanazawa2018end}. Another approach to deal with insufficient data is to use videos to leverage temporal information \cite{kocabas2020vibe,kanazawa2019learning}, along with generative-adversarial motion supervision. Another strategy involves concurrently utilizing 2D and 3D keypoints by iteratively improving the 3D mesh and measuring its projection error to 2D keypoints \cite{bogo2016keep}. SPIN \cite{kolotouros2019learning} utilizes this strategy to enhance the estimated outcomes within the training loop.

%\begin{figure}[ht!]
%    \includegraphics[width=0.48\textwidth]{Figures/SMPL_model.PNG}
%    \caption{The SMPL model \cite{SMPL}. The white points are pre-defined keypoints.}
%    \label{pic: SMPL model}
%\end{figure}

\subsection{3D Multi-Person Human Pose Estimation}
Multi-person methods can be categorized similarly to single-person methods. This means classification into skeleton and model-based approaches.
\subsubsection{Skeleton-based methods}
Similar to 2D multi-person HPE, the approaches can be classified into top-down and bottom-up methods. Top-down methods, such as LCR-net++ \cite{Lcr-net++} and the work by Moon et al. \cite{moon2019camera}, utilize anchor-based detection. On the other hand, following the bottom-up paradigm, Zehta et al. \cite{zanfir2018deep} employ single-image volumetric heatmaps for 3D keypoint prediction and confidence scores to group limbs belonging to the same individual. To address common inter-person occlusion challenges, an approach known as Occlusion-Robust Pose Maps \cite{mehta2018b} has been proposed. It incorporates redundant occlusion information into the part affinity maps.

\subsubsection{Model-based methods}
One of the most challenging tasks in model-based multi-person HPE is 3D mesh recovery. For instance, Zanfir et al. \cite{zanfir2018monocular} employ semantic segmentation and SMPL model fitting to tackle this challenge. In ROMP \cite{ROMP}, 2D heatmaps and mesh parameter maps are utilized for 2D body representation and 3D mesh localization. Building on this framework, Sun et al. \cite{BEV} introduce the Bird’s-Eye-View approach to infer depth from monocular cameras and categorize humans by age to employ different representations.

Moreover, video-based solutions integrating keypoint features, such as XNect \cite{XNect_SIGGRAPH2020}, provide real-time multi-person 3D human motion capture with just a single RGB camera. This algorithm effectively handles occlusion and introduces a novel CNN architecture featuring selective long and short-range skip connections for improved speed and accuracy. However, despite its effectiveness, XNect has limitations, including difficulty in handling neck occlusion, challenges in capturing extremely close interactions like hugging, and vulnerability of the identity tracker to drastic appearance changes and similar clothing.

\section{Datasets and Metrics}
\subsection{Datasets}
Datasets play a crucial role in fueling the advancement of deep learning methods for HPE. Their existence and accessibility enable the development, evaluation, and comparison of various algorithms. However, not all datasets are created equal; they vary significantly in terms of labels, annotations, and tasks they support.

In the realm of 2D HPE, standout datasets include the MPII Dataset \cite{MPII} and the MSCOCO Dataset \cite{COCO}. The MPII Dataset, sourced from YouTube videos, offers a diverse range of activities and annotations, including 16 2D keypoints, 3D torso keypoints, and occlusion labels. Similarly, the MSCOCO Dataset, compiled from webpage images, provides detailed annotations for 17 keypoints, along with bounding boxes and segmentation areas. Despite its focus on keypoint detection, MSCOCO supports additional tasks like object detection and panoptic segmentation. For multi-person HPE and tracking, researchers often rely on the PoseTrack Dataset \cite{posetrack}, known for its large-scale video-level annotations. This dataset has labels such as 15 2D keypoints, a person ID, and the head bounding box.

On the 3D HPE front, datasets exhibit a wider range of annotation types. For instance, the Human3.6M Dataset \cite{Human3.6M}, a single-person benchmark featuring annotations in 2D, 3D, and mesh formats. It contains 3.6 million human poses across various scenarios, captured using a sophisticated sensor setup comprising RGB cameras, a time-of-flight sensor, and motion cameras. Other notable datasets like HumanEva \cite{jain2010humaneva}, MPI-INF-3DHP \cite{mpi-inf-3d}, and MoVi \cite{movi} provide valuable poses of actors performing pre-defined actions, albeit with slight differences in action types and capture sensors across datasets. Such as 7 cameras for HumanEva, 14 cameras including 3 cameras with a top-down view for MPI-INF-3DHP, and IMU devices together with a motion capture camera for MoVi. 

Notably, MoVi stands out for its inclusion of mesh annotations alongside traditional 2D and 3D keypoints, providing a richer dataset for analysis. Detailed characteristics of individuals, such as age, height, and BMI are also supplied by this dataset. For all aforementioned datasets SMPL parameters can be obtained via the MoSh++ \cite{mosh}. Additionally, datasets like 3DPW \cite{3dpw} contribute to better model generalization in real-world scenarios by providing annotations of in-the-wild activities, albeit with limitations in crowded scenes. 3DPW utilizes IMUs to gather 2D, 3D, and mesh annotations in this challenging environment. Other important datasets include Leeds Sports Pose (LSP) Dataset \cite{leed_sport_pose_dataset}, Frames Labeled in Cinema (FLIC) Dataset \cite{sapp2013modec}, CrowdPose Dataset \cite{crowdpose}, J-HMDB Dataset \cite{jhuang2013towards}, CMU Panoptic \cite{joo2015panoptic}, AMASS \cite{mahmood2019amass}, and SURREAL \cite{varol2017learning}.

It's essential to acknowledge the limitations of existing datasets, particularly their constrained environments and limited pose diversity, as many datasets primarily capture poses involving standing or walking. Models trained on such datasets may struggle with generalization, especially for complex poses like those of gymnasts. Moreover, research has shown \cite{survey_HPE} that actors in datasets tend to move their upper bodies more than lower limbs, impacting model generalization, particularly for lower body parts.

\subsection{Metrics}
In 2D estimation, commonly used metrics include the Percentage of Correct Parts (PCP) or Keypoint (PCK), both employing an estimation threshold. PCK assesses each keypoint individually, while PCP focuses on the endpoints of a limb, ensuring they both fall within a threshold. Additionally, Average Precision (AP) and Recall (AR) are popular metrics for 2D estimation, typically evaluated at various Object Keypoint Similarity (OKS) values. These metrics are commonly presented individually for different OKS thresholds or summarized as the mean (mAP, and mAR respectively) of OKS ranging from 0.50 to 0.95 at intervals of 0.05. OKS addresses the limitation of Euclidean distance not generalizing well to different body shapes and sizes. It measures the distance between keypoints normalized by the person's scale, with distinct normalization values per keypoint to control fall-off. Similar normalization can also be used for PCP and PCK thresholds.

While the PCK metric is applicable in 3D HPE, the more widely used metric is the Mean Per Joint Position Error (MPJPE). MPJPE calculates the average Euclidean distance between ground truth and predictions, with poses aligned by the pelvis. Typically it is measured in millimeters. An alternative is the Procrustes Aligned MPJPE (PMPJPE), which mitigates the effects of translation, rotation, and scale through Procrustes alignment. Also known as the reconstruction error, PMPJPE places greater emphasis on reconstruction accuracy compared to MPJPE.
\section{SOTA Methods Overview and Comparison}
\subsection{2D Human Pose Estimation}
The SOTA methods for 2D and 3D HPE are presented in Tables \ref{table 2d} and \ref{table 3d}, respectively. In Table \ref{table 2d}, we compare the results of various 2D algorithms alongside their backbone architectures and input sizes. A notable observation from this table is the superior performance of top-down approaches over bottom-up approaches in multi-person HPE tasks. This discrepancy can be attributed to the inherent simplicity of top-down methodologies. However, recent trends suggest that bottom-up approaches may soon surpass their top-down counterparts. It is also noteworthy that tracking HPE algorithms exhibit inferior performance compared to those solely focusing on pose estimation. This phenomenon is observed in both top-down and bottom-up methods, respectively. The primary reason for this is the increased complexity associated with integrating tracking into pose estimation algorithms.

\begin{table}[htbp]
    \centering
    \addtolength{\leftskip} {-2cm} 
    \addtolength{\rightskip}{-2cm}
    \caption{Evaluation of representative 2D Human Pose Estimation methods and their performance. Single-person HPE algorithms were assessed on the MPII Test Set \cite{MPII}, while multi-person methods were evaluated on the COCO \cite{COCO}. Additionally, multi-person HPE and tracking algorithms underwent evaluation on the PoseTrack 2017 Test Set \cite{posetrack}. Notably, PCK represents PCKh@0.5, a popular PCK variant where the matching threshold is set to 50\% of the evaluated individual's head length.}
    \begin{tabular}{|c|c|c|>{\centering\arraybackslash}p{0.5cm}|c|}
        \hline
        Method & Backbone & Input size & PCK& mAP \\
        \hline\hline
        \multicolumn{5}{|c|}{Single-person HPE algorithms}\\
        \hline
        Tompson et al. \cite{tompson2014joint} & AlexNet & 320 $\times$ 240 & 79.6 &-\\
        Carreira et al. \cite{carreira2016human} & GoogleNet & 224 $\times$ 224 & 81.3&-\\
        Tompson et al. \cite{tompson2015efficient} & AlexNet & 256 $\times$ 256 & 82.0 &-\\
        Pishchulin et al. \cite{pishchulin2016deepcut} & VGG & 340 $\times$ 340 & 82.4 &-\\
        Gkioxary et al. \cite{gkioxari2016chained} & InceptionNet & 299 $\times$ 299 & 86.1 &-\\
        Insafutdinov et al. \cite{insafutdinov2016deepercut} & ResNet-152 & 340 $\times$ 340 & 88.5 &-\\
        Wei et al. \cite{wei2016convolutional} & CPM & 368 $\times$ 368 & 88.5 &-\\
        Newell et al. \cite{hourglass19} & Hourglass & 256 $\times$ 256 & 90.9 &-\\
        Li et al. \cite{li2021online} & Hourglass & 256 $\times$ 256 & 91.7 &-\\
        Chen et al. \cite{chen2017adversarial} & En/Decoder & 256 $\times$ 256 & 91.9 &-\\
        Sun et al. \cite{sun2019deep} & HRNet & 256 $\times$ 256 & 92.3 &-\\
        Tang et al. \cite{tang2019does} & Hourglass & 256 $\times$ 256 & \textbf{92.7} &-\\
        \hline
        \multicolumn{5}{|c|}{Multi-person bottom-up HPE algorithms}\\
        \hline
        OpenPose \cite{openpose} & CMU-Net & 368 $\times$ 368 &-& 61.8 \\
        Asso. Emb. \cite{newell2017associative} & Hourglass & 512 $\times$ 512 &-& 65.5 \\
        PifPaf \cite{pifpaf} & ResNet & 401 $\times$ 401 &-& 66.7 \\
        PersonLab \cite{personlab} & ResNet & 801 $\times$ 801 &-& 68.7 \\
        HigherHRNet \cite{cheng2020higherhrnet} & HRNet & 640 $\times$ 640 &-& 70.5 \\
        DEKR \cite{geng2021bottom} & HRNet & 640 $\times$ 640 &-& 71.0 \\
        SIMPLE \cite{zhang2021simple} & HRNet & 512 $\times$ 512 &-& 71.1 \\
        CenterGroup \cite{braso2021center} & HRNet & 512 $\times$ 512 &-& 71.4 \\
        SWAHR \cite{luo2021rethinking} & HRNet & 640 $\times$ 640 &-& \textbf{72.0} \\
        \hline
        \multicolumn{5}{|c|}{Multi-person top-down HPE algorithms}\\
        \hline
        G-RMI \cite{papandreou2017towards} & ResNe & 353 $\times$ 257 &-& 64.9 \\
        Integral Regre. \cite{sun2018integral} & ResNet & 256 $\times$ 256 &-& 67.8 \\
        CPN \cite{COCO_2017} & ResNet & 384 $\times$ 288 &-& 72.1 \\
        RMPE \cite{rmpe} & PyraNet & 320 $\times$ 256 &-& 72.3 \\
        SimpleBaseline \cite{xiao2018simple} & ResNet & 384 $\times$ 288 &-& 73.7 \\
        MSPN \cite{li2019rethinking} & MSPN & 384 $\times$ 288 &-& 76.1 \\
        DARK \cite{zhang2020distribution} & HRNet & 384 $\times$ 288 &-& 76.2 \\
        UDP [64] & HRNet & 384 $\times$ 288 &-& 76.5 \\
        PoseFix \cite{moon2019posefix} & HR+ResNet & 384 $\times$ 288 &-& 76.7 \\
        Graph-PCNN \cite{wang2020graph} & HRNet & 384 $\times$ 288 &-& 76.8 \\
        RSN \cite{cai2020learning} & 4-RSN & 384 $\times$ 288 &-& \textbf{78.6} \\
        \hline
        \multicolumn{5}{|c|}{Bottom-down multi-person HPE and tracking algorithms}\\
        \hline
        ArtTrack \cite{insafutdinov2017arttrack} &ResNet&256 $\times$ 256&-& 59.4\\
        PoseTrack \cite{posetrack} &ResNet&340 $\times$ 340&-& 59.4\\
        JointFlow \cite{doering2018jointflow} &SVG&256 $\times$ 192&-& 63.4\\
        STAF \cite{raaj2019efficient} &VGG& 368 $\times$ 368 &-& \textbf{70.3}\\
        \hline
        \multicolumn{5}{|c|}{Top-down multi-person HPE and tracking algorithms}\\
        \hline
        Detect-Track \cite{girdhar2018detect} &ResNet&256 $\times$ 256&-& 59.6 \\
        PoseFlow \cite{poseflow} &ResNet&-&-& 63.0\\
        PGPT \cite{bao2020pose} &ResNet&384 $\times$ 288&-& 72.6\\
        DetTrack \cite{wang2020combining} & HRNet &384 $\times$ 288&-& 74.1\\
        FlowTrack \cite{xiao2018simple} &ResNet&256 $\times$ 192&-& 74.6\\
        DCPose \cite{liu2021deep} &HRNet&384 $\times$ 288&-& \textbf{79.2}\\
        \hline
    \end{tabular}
    \label{table 2d}
\end{table}

Furthermore, it's evident that pioneering papers that introduced innovative approaches, like OpenPose \cite{openpose} or Associative Embeddings \cite{newell2017associative} in multi-person bottom-up HPE, often achieve lower scores compared to more refined methods that build upon them. These refined methods represent the current SOTA solutions. Among them, two notable papers are CenterGroup \cite{braso2021center} and SWAHR \cite{luo2021rethinking}. CenterGroup introduces an attention-based approach for multi-person HPE, focusing on accurately grouping keypoints based on attention mechanisms centred around keypoint centres. By employing attention mechanisms, the model dynamically weighs the importance of different keypoints relative to their spatial relationships, facilitating robust keypoint grouping. Experimental results demonstrate that the proposed method outperforms existing approaches, particularly in scenarios with crowded scenes and occlusions. On the other hand, SWAHR achieves even better results by rethinking the heatmap regression method. This novel strategy leverages both high-resolution and low-resolution representations to capture detailed pose information effectively. The proposed method achieves superior accuracy while maintaining computational efficiency. However, SWAHR's reliance on heatmap regression may introduce quantization errors during coordinate conversion.

For top-down approaches, which typically offer higher accuracy but slower processing times compared to bottom-up methods, PoseFix \cite{moon2019posefix} stands out for its exceptional performance. It is a model-agnostic pose refinement network aimed at improving the accuracy of existing HPE models. Unlike model-specific approaches, PoseFix is designed to refine the predictions of any pose estimator by learning to correct errors and inconsistencies in the initial estimations. By using both global context and local information, PoseFix effectively refines pose predictions, enhancing the overall quality of the estimated human poses. Nonetheless, its effectiveness hinges on the quality of the initial estimations; inaccurate or erroneous inputs may impede its ability to deliver optimal refinements. Conversely, Graph-PCNN \cite{wang2020graph} introduces a two-stage approach, leveraging a graph-based refinement stage after initial pose estimation, resulting in better-matched inputs for refinement. Thus it achieves even better results. However, the computational overhead of both methods might pose challenges, particularly in real-time applications, even despite their effectiveness.

Among the single-person HPE algorithms, Adversarial PoseNet \cite{chen2017adversarial} deserves some attention. It presents a novel CNN architecture that considers the structural relationships between body joints, resulting in more accurate pose estimations. Through adversarial training, the model generates realistic and diverse poses, enhancing its generalization performance. The network learns hierarchical features, capturing intricate pose details while maintaining robustness to scale variations and occlusions. However, its computational complexity might limit its applicability in resource-constrained or real-time settings. By the publication time, it was the SOTA model. However, further validation across diverse datasets and scenarios is necessary to fully understand its generalization and robustness. Later a better-performing method by Tang et al. \cite{tang2019does} emerged. The authors demonstrate that incorporating part-specific features enhances the performance, particularly in scenarios with occlusions and complex backgrounds. By focusing on related body parts, the proposed approach improves the model's ability to capture spatial dependencies and contextual information, leading to more accurate pose estimations.

While increased complexity often hampers accuracy in HPE methods with tracking, DCPose \cite{liu2021deep} challenges this norm by achieving superior results compared to other multi-person HPE algorithms. Its two-stage top-down architecture, featuring a coarse pose estimations network followed by a refinement network, contributes to its success. On the other hand, STAF \cite{raaj2019efficient} presents an intriguing bottom-up approach inspired by Openpose \cite{openpose}. It utilizes recurrent spatio-temporal affinity fields to model the associations between detections across frames. By iteratively refining the affinity fields using recurrent NN, the model effectively captures temporal dependencies while maintaining computational efficiency. While STAF demonstrates SOTA performance in accuracy and speed, its reliance on spatial information leads to sub-optimal initial accuracy, with optimal results achieved after longer frame sequences.

\subsection{3D Human Pose Estimation}
On the contrary, Table \ref{table 3d} illustrates that simpler 3D HPE techniques tend to outperform more intricate mesh-based 3D HPE methods. This observation is evident from the MPJPE and PMPJPE results on the Human3.6M dataset \cite{Human3.6M}. However, mesh-based algorithms demonstrate promising results and generalization capabilities on challenging in-the-wild datasets like 3DPW \cite{3dpw}. Considering these factors, along with the early stages of development for these models, substantial progress can be anticipated in this domain. Hence, these research directions hold significant promise for achieving notable advancements in addressing real-world challenges in HPE.

\begin{table}[htbp]
    \centering
    \addtolength{\leftskip} {-2.5cm} 
    \addtolength{\rightskip}{-2cm}
    \caption{Evaluation of representative 3D Human Pose, and Mesh Estimation methods on Human3.6M \cite{Human3.6M}, HumanEva-I \cite{jain2010humaneva}, 3DPW \cite{3dpw} datasets.}
    \begin{tabular}{|>{\centering\arraybackslash}p{2.4cm}|>{\centering\arraybackslash}p{0.8cm}|>{\centering\arraybackslash}p{0.85cm}|>{\centering\arraybackslash}p{1.5cm}|>{\centering\arraybackslash}p{0.8cm}|>{\centering\arraybackslash}p{0.85cm}|}
        \hline
        Method & \multicolumn{2}{|c|}{Human3.6M}& HumanEva-I & \multicolumn{2}{|c|}{3DPW}\\
        \hline
        & MPJPE & PMPJPE & PMPJPE & MPJPE &PMPJPE\\
        \hline\hline
        \multicolumn{6}{|c|}{3D HPE methods}\\
        \hline
        Zhou et al. \cite{20} & 113.0 & - & - & - &  -\\ 
        CHP \cite{CHP} & 92.4 & 59.1 & - & - &  - \\ 
Tome et al. \cite{tome2017lifting} & 88.4 & - & - & - &  -\\ 
C2F \cite{pavlakos2017coarse} & 71.9 & 51.9 & 25.5 & - &  -  \\ 

IHP \cite{sun2018integral} & 64.1 & 49.6 & - & - &  -  \\ 
Martinez et al. \cite{24} & 62.9 & 47.7 & 24.6 &  - &  - \\ 

SemGCN \cite{zhao2019semantic240} & 57.6 & - & - & - &  -  \\ 
Pavlakos et al. \cite{pavlakos2018ordinal} & 56.2 & 41.8 & 18.3 & - &  -  \\ 
3DMPPE \cite{moon2019camera} & 54.4 & - & - &  - &  - \\ 
Luvizon et al. \cite{luvizon20182d} & 53.2 & - & - &  - &  - \\ 
VideoPose3D \cite{pavllo20193d_156} & 46.8 & 36.5 & 19.7 &  - &  - \\
Xu et al. \cite{CHP_3D} & 45.6 & 36.2 & 15.2 & - &  -\\ 
Liu et al. \cite{liu2020attention} & 45.1 & 35.6 & 15.4 &  - &  -  \\ 
OANet \cite{planar_cylinder} & \textbf{42.9} & \textbf{32.8} & \textbf{14.3} & - &  -\\ 
        \hline
        \multicolumn{6}{|c|}{3D human mesh estimation methods }\\
        \hline
        SMPLify \cite{bogo2016keep} & 82.3 & - & - & - &  -\\ 
HMR \cite{kanazawa2018end} & 87.9 & 58.1& - & - &  -\\ 
Human dynamics \cite{kanazawa2019learning} & - & 56.9& - & - & 72.6 \\ 
GraphCMR \cite{kolotouros2019convolutional} & 71.9 & 50.1& - & - &  -\\ 
SPIN \cite{kolotouros2019learning} & - & 41.1& - & 96.9 & 59.2 \\ 
VIBE \cite{kocabas2020vibe} & 65.9 & 41.5& - & 93.5 & 56.5 \\ 
METRO \cite{lin2021end} & \textbf{54.0} & \textbf{36.7} & -& - &  -\\ 
SPEC \cite{kocabas2021spec} & - & - & -& - & \textbf{53.2} \\ 
ROMP \cite{ROMP} & - & - & -& \textbf{85.5} & 53.3 \\ 
\hline
    \end{tabular}
    \label{table 3d}
\end{table}

Some of the most interesting mesh methods are METRO \cite{lin2021end}, SPEC \cite{kocabas2021spec}, and ROMP \cite{ROMP}. METRO \cite{lin2021end} a single-person 3D HPE method proposes an end-to-end framework for simultaneously reconstructing 3D human poses and meshes from single images using transformers. Unlike traditional approaches that rely on separate stages for pose and mesh estimation, this method integrates both tasks into a unified architecture. The framework leverages a transformer-based backbone to encode the input image and generate intermediate pose and mesh representations. These representations are then refined using transformer decoder layers to produce the final 3D pose and mesh outputs. By jointly optimizing both tasks in an end-to-end manner, the proposed method achieves the smallest MPJPE and PMPJPE errors on the Human3.6M \cite{Human3.6M} dataset among all 3D mesh-based approaches. An alternative approach, SPEC \cite{kocabas2021spec}, is specifically designed for handling in-the-wild HPE. By integrating camera parameters like focal length, SPEC enhances the geometric coherence of pose predictions in challenging outdoor environments. This approach enables more robust pose estimation even in scenarios with varying camera viewpoints and distances. Furthermore, SPEC introduces a novel dataset and evaluation protocol, providing a benchmark for assessing human pose estimation methods in outdoor settings. The dataset encompasses images captured across diverse outdoor scenes, spanning streets, parks, and public areas, with varying occlusion levels, lighting conditions, and background complexities. Additionally, it provides annotated ground truth poses and corresponding camera parameters, facilitating method evaluation. SPEC's evaluation framework includes conventional metrics like accuracy, precision, and recall, alongside new criteria assessing pose estimation robustness to changes in camera perspective and distance. While SPEC addresses some of the requirements for tailored datasets and metrics in mesh-based 3D HPE, the ultimate efficacy and community acceptance of their approach remains uncertain. Consequently, there remains substantial room for improvement and further discoveries in this field. Another compelling multi-person solution is ROMP \cite{ROMP}. This paper introduces a monocular, one-stage regression approach for estimating multiple 3D human body meshes from a single RGB image. The model leverages a regression-based architecture that directly predicts the 3D joint locations of each person in the image without necessitating intermediate steps or bounding boxes. The method entails predicting a body centre heatmap and a mesh parameter map, which intricately describe 3D body meshes at a pixel level, and subsequently extracting body mesh parameters via a body-centre-guided sampling process. Experimental findings on challenging multi-person benchmarks showcase ROMP's superior performance over SOTA, particularly in crowded and occluded environments. It is achieved by the Collision-Aware Representation that adeptly tackles centre ambiguity in crowded scenarios. Notably, ROMP is the first real-time implementation of monocular multi-person 3D mesh regression. However, ROMP's performance might degrade in novel environments or with significantly different conditions from its training data.

Some of the top-performing 3D HPE methods leverage temporal video data for refining predictions. One such method, OANet \cite{planar_cylinder}, addresses occlusion challenges by integrating occlusion-aware modules and temporal convolutional networks into its architecture, thus leveraging both spatial and temporal information. Similar accuracy is achieved by Liu et al. \cite{liu2020attention}, as they propose to use attention mechanisms to exploit temporal contexts. The method utilizes a recurrent NN architecture augmented with attention mechanisms to capture temporal dependencies in sequential data. By dynamically weighting the importance of temporal contexts, the model enhances its ability to predict 3D human poses in real-time from streaming video inputs. However, there are also notable methods focusing on single images. For instance, Xu's method \cite{CHP_3D} enhances accuracy and robustness by incorporating deep kinematics analysis. \cite{CHP_3D} integrates kinematic constraints into the network architecture to infer more plausible 3D human poses while considering anatomical constraints and joint relationships. This work extends the 2D CHP \cite{CHP} approach to 3D.

\section{Future Work}
The field of 3D HPE and Modeling continues to grapple with several unresolved challenges. One such challenge is the estimation of complex postures, such as those exhibited by athletes like gymnasts. Addressing this challenge may involve creating novel datasets tailored to specific complex and rare postures or exploring the use of unsupervised or generative models.

Crowded scenes present another significant problem, particularly challenging for 3D reconstruction \cite{cheng2020higherhrnet,crowdpose}. In such scenarios, occlusion and interperson interactions occur often, complicating the estimation process. These scenarios, along with person-object interactions, pose another challenge for researchers. Although some progress has been made in modelling hand-object interactions \cite{tekin2019h+} and body interactions with specific objects \cite{hassan2019resolving}, we are still far from achieving systems accurately operating in complex real-world environments.

Realistically modelling the entire human body, including fine-grained details of hand movement and facial appearance, presents another formidable challenge. While frameworks for whole-body representation have been proposed \cite{pavlakos2019expressive, xiang2019monocular}, they often fall short of capturing intricate face and hand details. Addressing this challenge may involve creating datasets tailored to fine-grained whole-body representation and developing novel unsupervised or weakly supervised models. Furthermore, cutting-edge computer graphics solutions like NeRF \cite{nerf} and gaussian splatting \cite{gaussian_splatting} offer promising avenues for modelling fine-grained facial details. Additionally, there is room for improvement in modelling facial expressions, emotional states, and dynamic movement of clothes. Success in this area could pave the way for virtual digital human applications such as real-time telepresence, virtual customer service, improved computer-generated scenes in movies, and virtual reality. These require combining detailed representations of the entire human body with accurate portrayals of appearance, gender, and personality. It also involves capturing nuances such as lip movement during speech, as well as the ability to communicate using language, emotions, body language, and facial expressions through digital humans.

Furthermore, developing better benchmarks for mesh-based 3D human body reconstruction is crucial. Currently, we primarily rely on skeleton-based metrics like MPJPE and PMPJPE, which do not fully account for the complexities of mesh representations, including appearance. Additionally, with the increased complexity of 3D mesh models, there is a pressing need for extensive datasets to improve model training. Finally, transitioning 3D mesh modelling from research to industry will require the development of user-friendly toolkits and programming interfaces.

\section{Conclusion}
In conclusion, we've covered the essential background for understanding human modelling and pose estimation in both 2D and 3D domains. We introduced popular datasets, metrics, and state-of-the-art algorithms, comparing their strengths and weaknesses. Furthermore, we explored potential future research directions, all aimed at advancing towards a common goal. Given that many current approaches are task-specific, there is a pressing need to prioritize the development of more general human representations, particularly in natural, real-world environments.

\bibliographystyle{IEEEtran}
\bibliography{main}

\begin{IEEEbiography}[{\includegraphics[width=1in,height=1.25in,clip,keepaspectratio]{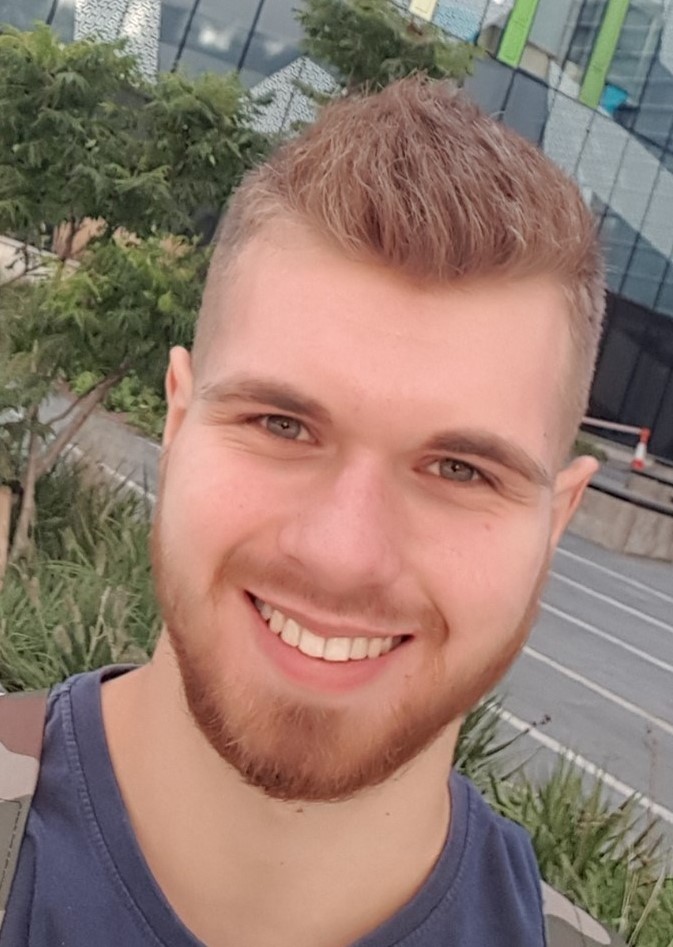}}]{Pawel Knap} is a PhD student at the University of Freiburg, Germany. He completed his MEng in Electronic Engineering with Artificial Intelligence at the University of Southampton. Pawel is the first author of two peer-reviewed conference papers. His academic interests primarily focus on computer vision, particularly using AI methods for medical image data analysis. For more about Pawel please visit: pawelknap.github.io
\end{IEEEbiography}
\end{document}